\documentclass[pdflatex,sn-mathphys-num]{sn-jnl}% Math and Physical Sciences Numbered Reference Style
\usepackage{graphicx}%
\usepackage{multirow}%
\usepackage{amsmath,amssymb,amsfonts}%
\usepackage{amsthm}%
\usepackage{mathrsfs}%
\usepackage[title]{appendix}%
\usepackage{xcolor}%
\usepackage{textcomp}%
\usepackage{manyfoot}%
\usepackage{booktabs}%
\usepackage{algorithm}%
\usepackage{algorithmicx}%
\usepackage{algpseudocode}%
\usepackage{listings}%
\usepackage{subcaption}

\usepackage[final]{changes}

\theoremstyle{thmstyleone}%
\theoremstyle{thmstyletwo}%

\theoremstyle{thmstylethree}%

\newcommand{\tablestd}[1]{{\footnotesize$\pm$#1}}

\begin{document}

\title[Article Title]{Towards Comprehensive Real-Time Scene Understanding in Ophthalmic Surgery through Multimodal Image Fusion}

%%=============================================================%%
%% GivenName	-> \fnm{Joergen W.}
%% Particle	-> \spfx{van der} -> surname prefix
%% FamilyName	-> \sur{Ploeg}
%% Suffix	-> \sfx{IV}
%% \author*[1,2]{\fnm{Joergen W.} \spfx{van der} \sur{Ploeg} 
%%  \sfx{IV}}\email{iauthor@gmail.com}
%%=============================================================%%

\author*[1,2]{\fnm{Nikolo} \sur{Rohrmoser}}\email{nikolo.rohrmoser@tum.de}

\author[2]{\fnm{Ghazal} \sur{Ghazaei}} %\email{ghazal.ghazaei@zeiss.com}
\equalcont{These authors contributed equally to this work.}

\author[1]{\fnm{Michael} \sur{Sommersperger}}%\email{michael.sommersperger@tum.de}
\equalcont{These authors contributed equally to this work.}

\author[1]{\fnm{Nassir} \sur{Navab}}%\email{nassir.navab@tum.de}

\affil[1]{\orgdiv{Chair for Computer Aided Medical Procedures}, \orgname{Technical University of Munich}, \orgaddress{\city{Garching}, \country{Germany}}}

\affil[2]{\orgdiv{Corporate Research and Technology}, \orgname{Carl Zeiss AG}, \orgaddress{\city{Munich}, \country{Germany}}}

%%==================================%%
%% Sample for unstructured abstract %%
%%==================================%%

% \abstract{\textbf{Purpose:} The abstract serves both as a general introduction to the topic and as a brief, non-technical summary of the main results and their implications. The abstract must not include subheadings (unless expressly permitted in the journal's Instructions to Authors), equations or citations. As a guide the abstract should not exceed 200 words. Most journals do not set a hard limit however authors are advised to check the author instructions for the journal they are submitting to.}

%%================================%%
%% Sample for structured abstract %%
%%================================%%

\abstract{ % 250 words limit
\textbf{Purpose:}
The integration of multimodal imaging into operating rooms paves the way for comprehensive surgical scene understanding.
In ophthalmic surgery, by now, two complementary imaging modalities are available: operating microscope (OPMI) imaging and real-time intraoperative optical coherence tomography (iOCT).
This first work toward temporal OPMI and iOCT feature fusion demonstrates the potential of multimodal image processing for multi-head prediction through the example of precise instrument tracking in vitreoretinal surgery.

\textbf{Methods:} We propose a multimodal, temporal, real-time capable network architecture to perform joint instrument detection, keypoint localization, and tool-tissue distance estimation.
Our network design integrates a cross-attention fusion module to merge OPMI and iOCT image features, which are efficiently extracted via a YoloNAS and a CNN encoder, respectively.
Furthermore, a region-based recurrent module leverages temporal coherence.

\textbf{Results:} Our experiments demonstrate reliable instrument localization and keypoint detection (95.79\% mAP50) and show that the incorporation of iOCT significantly improves tool-tissue distance estimation, while achieving real-time processing rates of 22.5 ms per frame.
Especially for close distances to the retina (below 1 mm), the distance estimation accuracy improved from 284 µm (OPMI only) to 33 µm (multimodal).

\textbf{Conclusion:}
Feature fusion of multimodal imaging can enhance multi-task prediction accuracy compared to single-modality processing and real-time processing performance can be achieved through tailored network design. While our results demonstrate the potential of multi-modal processing for image-guided vitreoretinal surgery, they also underline key challenges that motivate future research toward more reliable, consistent, and comprehensive surgical scene understanding.

}


%%\pacs[JEL Classification]{D8, H51}

%%\pacs[MSC Classification]{35A01, 65L10, 65L12, 65L20, 65L70}

\maketitle
\section{Introduction}

\begin{figure}[h!tb]
    \centering
    \includegraphics[width=\linewidth]{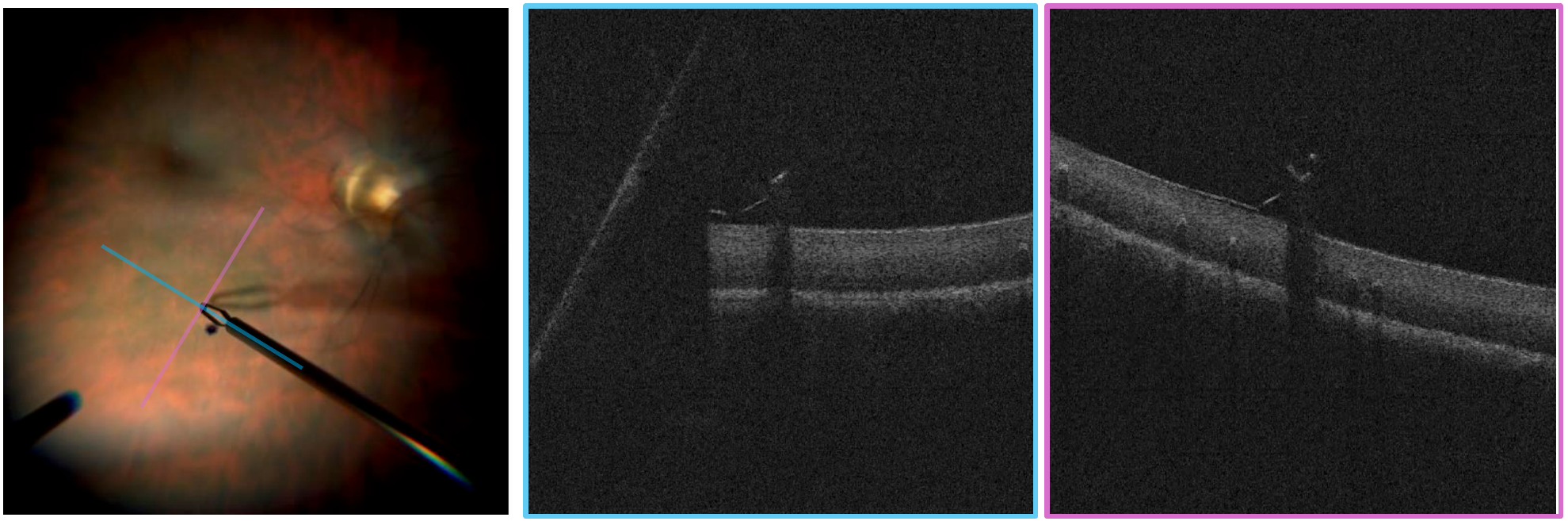}
    \caption{Multimodal data in ophthalmic surgery consists of OPMI (left) and two perpendicularly aligned iOCT B-scan images (middle, right). The blue and magenta lines in the OPMI image indicate the locations of the B-scans.}
    \label{fig:MultiModalData}
\end{figure}

Ophthalmic, and in particular vitreoretinal surgeries, are among the most intricate minimally invasive surgical procedures, as they involve the precise manipulation of delicate ocular tissue structures.
Ensuring safety during delicate tool–tissue interactions is essential to prevent irreversible retinal damage, with surgical visualization being central to achieving precise and reliable outcomes.
Traditionally, vitreoretinal surgeons solely rely on the view through the operating microscope (OPMI).
While this view provides a wide-angle visualization of the retina, it does not allow for micrometer resolution imaging of subsurface layer structures, subtle retinal pathologies, and precise estimation of the instrument position relative to the ocular anatomy.
These capabilities have become available within the last decade with the integration of optical coherence tomography (OCT) into surgical microscopes, providing additional real-time cross-sectional imaging with micrometer-level axial resolution.
The multimodal data available in current setups, hence, comprises OPMI imaging, each frame paired with a set of two perpendicular cross-sectional intraoperative OCT (iOCT) B-scans, as shown in Fig. \ref{fig:MultiModalData}.
Studies have shown that iOCT can enhance the identification and assessment of anatomical and pathological tissues for surgical decision-making \cite{ehlers2018outcomes,posarelli2020impact} and improve the precision of depth-related instrument targeting \cite{schutz2025impact}.

Beyond direct surgical visualization, OPMI and iOCT imaging have been employed for many advanced image-guided vitreoretinal applications.
In manual surgery, real-time image processing of iOCT enables the visual \cite{roodaki2015introducing} and auditive \cite{matinfar2024ocular} augmentation of the intraoperative environment to support the surgeon during complex instrument maneuvers.
On the other hand, microscope imaging with its wide field of view allows capturing the surgical context and has been used for instrument tracking \cite{rieke2016real}, surgical phase understanding \cite{koksal2024sangria,gastager2025watch}, and autonomous navigation of the instrument to the retina in robotic surgery, relying on the optical shadow, which the instrument casts on the retina, for distance estimation \cite{kim2020autonomously,yang2024shadowpose}.
Moreover, iOCT has been widely investigated for image-guided robotic surgery.
In recent works, it was shown that real-time image processing of iOCT could even enable eventual autonomous robotic procedures, such as for needle insertion and subretinal injection \cite{arikan2025real}, achieving the required precision for such complex procedures.

Integrating information from multiple imaging modalities enables a more comprehensive understanding of the surgical environment and supports more precise and reliable model predictions.
However, most existing methods remain confined to single-modality inputs, lacking unified architectures capable of reasoning across complementary data streams to achieve holistic surgical scene understanding and overcome the inherent limitations of analyzing only a single modality.
Fusing information from multimodal data could improve the robustness and precision of learning-based approaches and even enable new applications in image-guided ophthalmic surgery.

Advances in computer vision have shown that fusing complementary modalities enhances both spatial and temporal understanding. In static perception, RGB-D fusion couples appearance and geometry, with architectures such as RDFNet \cite{park2017rdfnet} and DenseFusion \cite{wang2019densefusion} demonstrating how depth cues refine segmentation, localization, and 6-DoF pose estimation. For dynamic scenes, two-stream models \cite{simonyan2014two,carreira2017quo} integrate optical flow and RGB information to jointly encode motion and structure. Beyond these, the inclusion of thermal, LiDAR, and event-based signals \cite{zhang2023cmx} has further extended perception under challenging or high-speed conditions.
In the surgical domain, Laina et al. \cite{laina2017concurrent} combined depth and appearance for accurate instrument tracking, while multimodal transformers \cite{weerasinghe2024multimodal} integrated visual and kinematic streams for real-time recognition of fine-grained surgical activities. The Surgical Flow Masked Autoencoder \cite{mostafa2025surgical} later leveraged optical-flow priors to model motion-driven events. These advances establish multimodal learning as a central paradigm for improving robustness, spatial awareness, and temporal reasoning laying the foundation for OCT–microscope fusion in ophthalmic surgeries \cite{wang2024advances}.

In this work, we present the first multimodal (MM) and recurrent multimodal (RMM) framework for temporal OPMI and iOCT feature fusion, enabling joint keypoint detection, classification, and precise estimation of the instrument-to-retina distance.
To fully exploit the complementary strengths of both modalities, we design a cross-attention fusion module within the Yolo-NAS framework \cite{supergradients}, which merges high-level representations from its OPMI encoder with structural depth cues from a dedicated iOCT extractor. A recurrent extension further enhances temporal coherence and robustness of the fused representation, ensuring stable predictions over time.
The resulting architecture unifies the advantages of both modalities into a shared feature space, supporting efficient multi-head prediction across key surgical understanding tasks.

To demonstrate the potential of our framework, we perform experiments on a photorealistic synthetic multimodal dataset, given the current lack of annotated temporal multimodal datasets of the desired vitreoretinal scenarios.
Our experiments demonstrate that the proposed fusion approach effectively integrates complementary visual and depth information, achieving high instrument detection accuracy (95.79\% mAP50) and substantially improving tool–tissue distance estimation compared to single modality (SM) processing.
Most notably, for close-range interaction, the error decreases from 284 µm (single modality) to just 33 µm (multimodal), underlining the potential of multimodal fusion for safer, depth-aware surgical guidance.
Finally, we examine the reliability and interpretability of the fusion mechanism and discuss current limitations and future research directions toward comprehensive multimodal scene understanding.

\section{Methodology}
\label{sec:methodology}

\begin{figure}[t]
    \centering
    \includegraphics[width=0.9\textwidth]{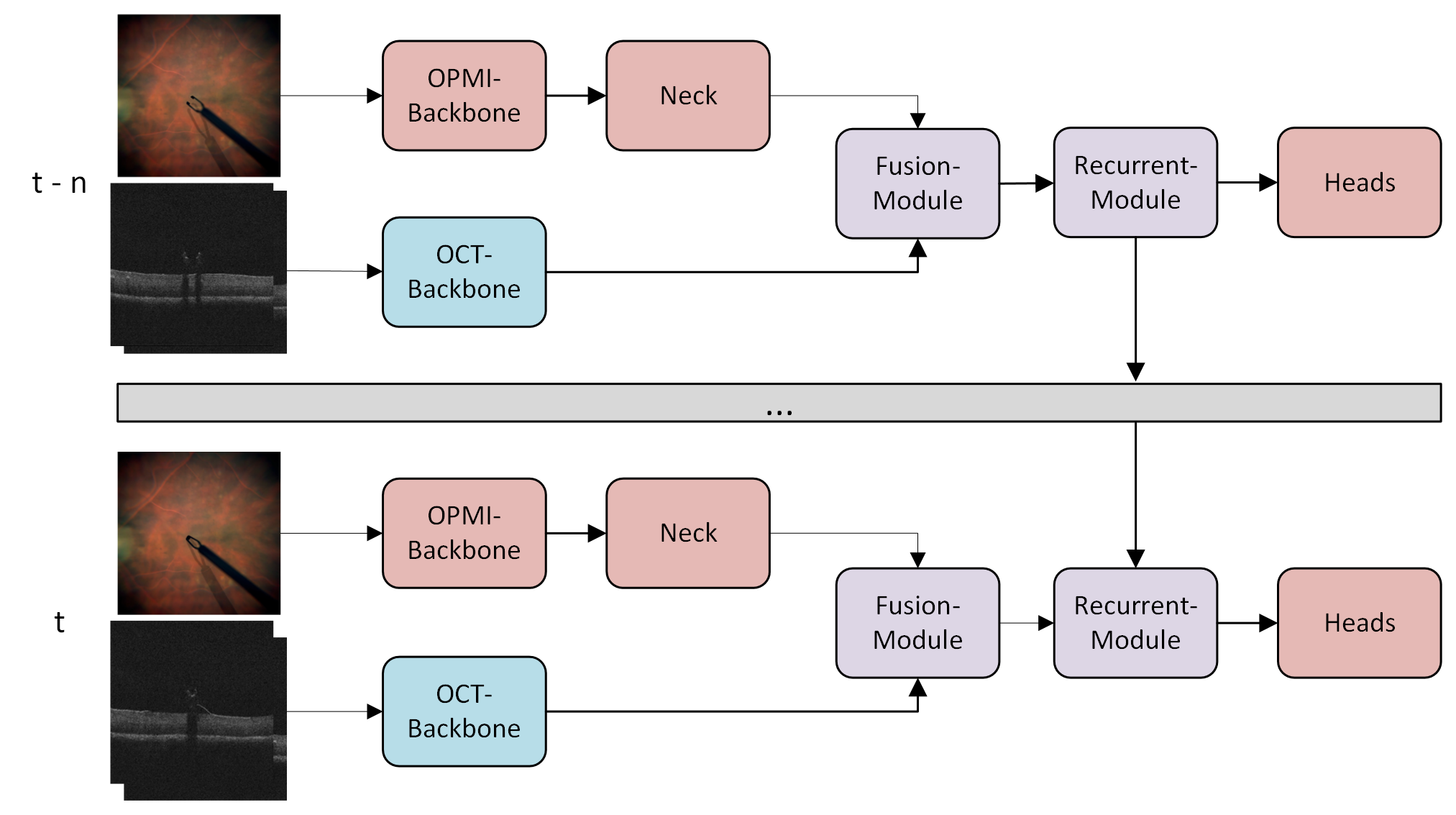}
    \caption{The overall architecture. A Yolo-NAS backbone processes the OPMI stream, while a modified ResNet-18 extracts features from iOCT scans. An attention-based fusion module integrates these modalities before they are further refined by a recurrent module for temporal awareness. Finally, task-specific heads generate outputs including detection, keypoint estimation, and distance prediction.
    }
    \label{fig:04_architecture}
\end{figure}

We propose a multimodal, temporally consistent architecture that extends a single-modality one-stage detector with a secondary branch for multimodal data integration, a cross-modal attention fusion module, and a lightweight recurrent unit for temporal refinement. The framework jointly performs instrument detection, keypoint localization, and tool–tissue distance estimation.
Illustrated in Figure~\ref{fig:04_architecture}, the network structure comprises two parallel streams.
The primary stream uses a Yolo-NAS-m backbone to process OPMI frames, generating multi-scale feature pyramids that capture hierarchical spatial context across the surgical scene. We call this base architecture Single-Modality (SM) model from now on.
To integrate iOCT, the secondary stream employs a modified ResNet-18 encoder, extended with an adaptive pooling layer that preserves lateral resolution. 
This design extracts $M$ column descriptors of size $C_{OCT}$, each representing neighboring A-scans, maintaining the spatial correspondence between the depth profiles (iOCT features) and their position within the microscopic view.

\subsection{Fusion Module}
\label{subsec:fusion}

\begin{figure}[t]
    \centering
    \includegraphics[width=0.7\textwidth]{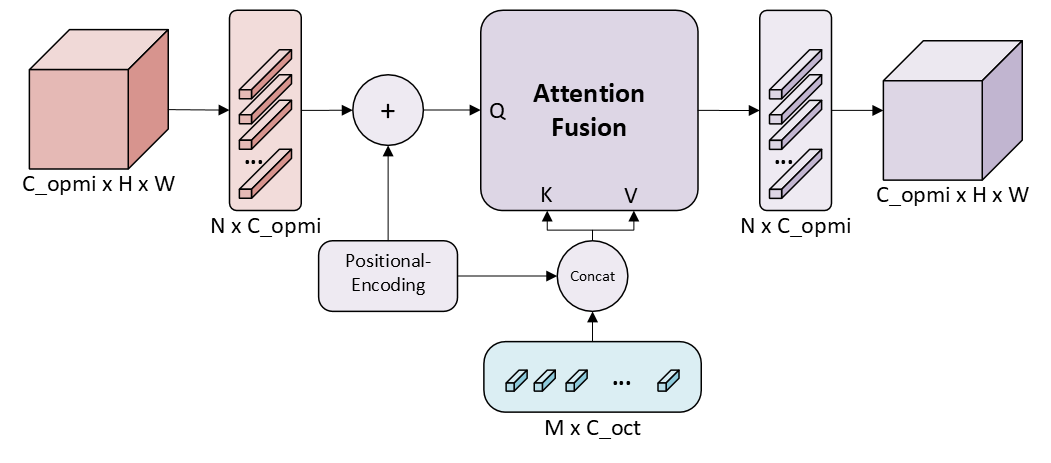}
    \caption{Cross-attention fusion. OPMI pixel features (queries) attend to iOCT column descriptors (keys/values) enriched with positional encodings to inject spatially aware depth cues into the fused representation. The output retains the spatial resolution and channel dimensionality required by the downstream prediction heads.}
    \label{fig:04_attn_fusion}
\end{figure}

The core of our multimodal integration is the attention-based fusion module applied independently to feature maps of different resolutions produced by the Yolo-NAS neck.
Fusing these modalities presents distinct geometric challenges: OPMI provides an en-face surface view, while iOCT captures orthogonal cross-sectional depth. This discrepancy makes direct pixel-level alignment ill-posed. Furthermore, depth information is spatially sparse and only relevant to specific instrument locations, requiring a mechanism that avoids global feature mixing. Therefore, we propose a streamlined design to effectively address this alignment problem by enabling OPMI features to selectively query iOCT data based on semantic relevance, rather than relying on fixed spatial correspondence. 
Consequently, as illustrated in Figure~\ref{fig:04_attn_fusion}, this module employs a cross-attention mechanism to dynamically inject the iOCT-derived depth cues into the OPMI feature maps.
The attention-fusion block consists of stacked multi-head attention layers followed by a feed-forward MLP, similar to the definition of a transformer encoder block~\cite{vaswani2017attention}. Unlike the standard encoder, which relies solely on self-attention, our formulation performs cross-attention by treating our OPMI pixel features as queries and the iOCT features as keys and values. In our framework, two such blocks are stacked to provide a trade-off between expressiveness and computational efficiency.
Sinusoidal positional encodings are added to the OPMI queries, while each iOCT column descriptor is concatenated with a positional code derived from its projected location in the OPMI grid. This enables the network to model the spatial correspondence between modalities and to localize the iOCT imaging region within the microscope view. We refer to this architecture Multimodality (MM) model from now on.

\subsection{Recurrent Module}
\label{subsec:recurrent_module}
To enhance temporal consistency and increase robustness to transient data corruption, we integrate a recurrent module following the feature fusion stage.
The fused feature map is first downsampled using average pooling to generate a compact grid of regional descriptors, reducing computational overhead. A recurrent unit then processes the temporal sequence of these descriptors across consecutive frames. The resulting temporally-aware features are upsampled via interpolation to the original feature map resolution and combined with the fused representation. We refer to the final model as Recurrent Multi Modality (RMM) model.
The recurrent module is supervised by the cosine contrastive loss, encouraging feature similarity between representations derived from corrupted ($f_c$) and intact ($f_n$) iOCT inputs ($\mathcal{L}_{\text{cos}} = 1 - \cos(f_c, f_n)$).

\subsection{Prediction Heads}
\label{subsec:distance_head}
Yolo-NAS prediction heads operate independently on the multi-scale feature maps produced by the model’s neck.
In our framework, the heads receive the temporally refined, fused features, and each is supervised by its own task-specific loss function.
The Yolo-NAS loss formulation includes separate components for keypoint visibility (binary cross-entropy), keypoint regression (smooth L1), object classification (focal loss), and bounding-box regression (distribution focal loss and GIoU-based loss)~\cite{supergradients}.
More details on the loss functions are provided in the supplementary material.
All loss components are combined in a weighted sum for end-to-end training.
Each head produces predictions at every spatial location of the feature map, which are subsequently decoded into final detections following the standard Yolo-NAS post-processing pipeline.
This modular design preserves the original Yolo-NAS detection and keypoint heads while allowing the integration of additional, task-specific output heads. 

To integrate a new head for tool–tissue distance estimation, we adopt the distributional regression formulation used for bounding-box regression in Yolo-NAS.
Instead of regressing a single scalar value, the head predicts a discrete probability distribution over distance bins.
It consists of two convolutional blocks ($1 \times 1$ and $3 \times 3$ kernels) with batch normalization and ReLU activation, followed by a $1 \times 1$ convolution producing $\text{reg}_{\max} + 1$ logits.
These logits are converted into a probability distribution $\mathbf{p}=(p_0,\ldots,p_{\text{reg}_{\max}})$ via softmax, yielding the final distance prediction:
\begin{equation}
    \hat{y} = \sum_{i=0}^{\text{reg}_{\max}} i\, p_i
\end{equation}
The expected value $\hat{y}$ is linearly scaled from the bin index space to metric units (e.g., mm) using the known application-specific depth range $[d_{\min}, d_{\max}]$.
Similar to Yolo-NAS's bounding box regression, the training of this head is supervised using a distribution-focal loss \cite{li2020generalizedfocallossv2}, which encourages the model to learn a sharp probability distribution centered around the ground-truth distance.

A key advantage of this formulation is that it allows deriving a certainty score directly from the predicted distribution, as the sharpness correlates with prediction accuracy \cite{li2020generalizedfocallossv2}.
For the most likely bin $i^* = \arg\max_i p_i$, we compute the certainty as
$p_{i^*} + \max\big( p_{i^*-1},\; p_{i^*+1} \big)$.

\section{Experiments}
\label{sec:experiments}

\bmhead{Data}\label{subsec:data}
Although iOCT-integrated surgical microscopes have become commercially available, specialized multimodal datasets with specific imaging conditions, such as aligning the iOCT with the instrument, are not yet available.
Acquiring such specific datasets in clinical settings requires precise registration between iOCT and micropscope imaging, as well as instrument tracking to correctly position the B-scans along and perpendicular to the instrument axis. 
While these aspects are technically feasible, and multimodal alignment \cite{turgut2025real} and instrument tracking \cite{rieke2016real} have been addressed separately by the research community, they have not yet been incorporated into commercial devices, making the generation of datasets for studies such as this extremely challenging. Synthetic data, on the other hand, enables the exploration of advanced sensor fusion methods without being limited by the capabilities of currently available commercial systems.
In addition, the full annotation of multimodal temporal data is highly labor-intensive.
In this initial work, to overcome the limitations of currently available devices and to avoid laborious expert annotations, we present a systematic and controlled evaluation on curated data.
The synthetic dataset\footnote{SynthesEyes GmbH, Munich, Germany} contains photorealistic, fully annotated, and synchronized multimodal OPMI and iOCT data, comprising 20 videos (69,134 frames, 45-60 FPS) from two types of simulated surgeries, including retinal membrane peeling and subretinal injection, featuring four instrument classes (two types of forceps, a surgical cannula, and an endoilluminator). 
Each video includes OPMI and two orthogonal iOCT B-scans centered at the tooltip, with one B-scan aligned with the instrument, as shown in Fig. \ref{fig:MultiModalData}.
The synthetic iOCT data is modeled closely after real-world OCT data, incorporating typical iOCT artifacts, such as instrument shadowing that obscures the retinal anatomy below the instrument and mirroring along the OCT zero-delay, arising from the complex conjugate extracted during signal reconstruction in real-world spectral and swept-source OCT systems.
The dataset provides full ground truth information, including instrument keypoints, multimodal segmentation masks for surgical instruments and retinal layer and vessel structures, and instrument-to-retina distances. The data is split into 17:3 videos for training and validation. The resolution of each input is resized to 512x512 pixels.
The videos are subsampled, using every 30th frame for frame-wise predictions, and every 10th frame for temporal experiments, to avoid near-duplicate samples, as instruments are often carefully navigated at a slow speed during delicate retinal procedures.
Both modalities undergo normalization and augmentation with moderate color, noise, and fog variations to simulate realistic variations of imaging conditions.

\bmhead{Experimentation Setup}\label{subsec:experimentats}
As a basis for all models, we use the Yolo-NAS-m implementation from the SuperGradients library~\cite{supergradients} with COCO-pretrained weights~\cite{lin2015coco}. We train for 50 epochs (120 epochs for temporal experiments) using the AdamW optimizer with a cosine learning rate schedule
and a learning rate of 1.2726e-4, where hyperparameters (including loss weights and the learning rate) were determined via an Optuna search jointly optimizing mAP50 and dMAE.
All experiments were conducted on a single NVIDIA GeForce RTX 3090.
In our configuration, the iOCT-Embedder generates $M=16$ column descriptors, each with a dimensionality of $C_{OCT}=16$ per B-scan. This follows from the input resolution and the downsampling properties of the employed ResNet-18 backbone. For distance estimation, we define a depth range of $d_{\min}=-1$ and $d_{\max}=6mm$.
For temporal experiments, the RMM is trained with sequences of 16 consecutive frames. During training, the outputs of the iOCT embedder for 4 consecutive and 4 randomly selected frames are replaced with randomized iOCT feature vectors to simulate unreliable iOCT data. During inference, we evaluate the model's performance for a test video with varying numbers of corrupted iOCT frames using the same randomization strategy as above.

\bmhead{Metrics}\label{subsec:metrics}
The models are evaluated on three tasks: instrument detection, keypoint localization, and instrument-to-retina distance estimation. Detection performance is measured using mean average precision at 50\% IoU (mAP50). Keypoint localization is evaluated by the average pixel distance between predicted and ground-truth keypoints (kp dist). Distance estimation accuracy is assessed using mean absolute error (dMAE) in micrometers, evaluating both the overall distances and specifically distances below 1 mm (dMAE0:1). Additionally, we report dMAE restricted to predictions with high certainty (\textgreater90\%) to evaluate the effectiveness of the certainty measure.

\section{Results \& Discussion}
\label{subsec:results}

\begin{table}[h!tb]
    \caption{Comparison of SM and MM models on detection, keypoint localization, and distance estimation on single frames without utilization of the recurrent module. The metrics for the RMM models are reported using a sequence of 16 frames and 4 consecutive and 4 random corrupted iOCT features using a GRU/LSTM as recurrent module.}
        \label{tab:sm_mm_rmm}
        \centering
    \begin{tabular}{lrrrrrr}
        \toprule
        \multirow{2}{*}{Model}
        & mAP50 & kp dist & dMAE & dMAE0:1 & certain dMAE \\
        & [\%] & [pxl] & [$\mu m$] & [$\mu m$] & [$\mu m$] \\
        \midrule
        SM & 94.27  \tablestd{0.41} & 8.35 \tablestd{1.32} & 480.93 \tablestd{54.91} & 284.01 \tablestd{89.44} & 142.99	\tablestd{25.4} \\
        MM & 95.79 \tablestd{0.58} & 7.93 \tablestd{0.47} & 128.32 \tablestd{20.03} & 33.05 \tablestd{2.38} & 53.74 \tablestd{13.0} \\
        \midrule
        RMM (GRU)& 94.92 \tablestd{0.28} & 9.65 \tablestd{0.42} & 332.91 \tablestd{48.88} & 135.44 \tablestd{37.28} & 61.78 \tablestd{10.73} \\
        RMM (LSTM) & 94.60 \tablestd{0.29} & 9.17 \tablestd{0.46} & 351.59 \tablestd{34.66} & 141.85 \tablestd{67.60} & 62.88 \tablestd{14.54} \\
        \bottomrule
    \end{tabular}
\end{table}

\begin{figure}[h!tb]
    \centering
    \includegraphics[width=0.8\textwidth]{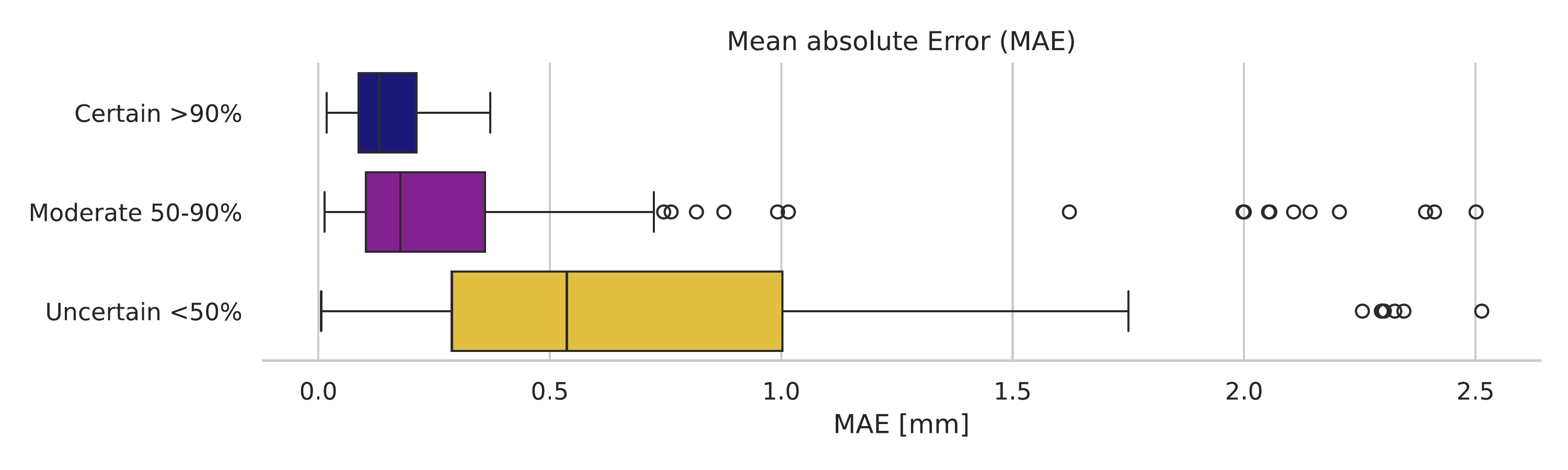}
    \caption{Distance error distributions for a single peeling sequence in the test set stratified by certainty for the SM model utilizing only OPMI. Each boxplot summarizes the error distribution for one certainty category of certain (\textgreater90\%), moderate (50-90\%), and uncertain (\textless50\%).}
    \label{fig:certainty_vs_MAE}
\end{figure}

\begin{figure}[h!tb]
    \centering
    \begin{subfigure}[b]{.9\textwidth}
        \includegraphics[width=\textwidth]{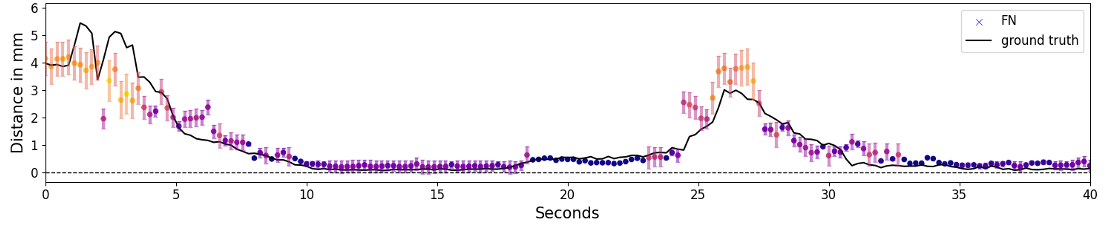}%
        \caption{Single Modality}
        \label{fig:sm_signal}
    \end{subfigure}
    \vfill
    \begin{subfigure}[b]{.9\textwidth}
        \includegraphics[width=\textwidth]{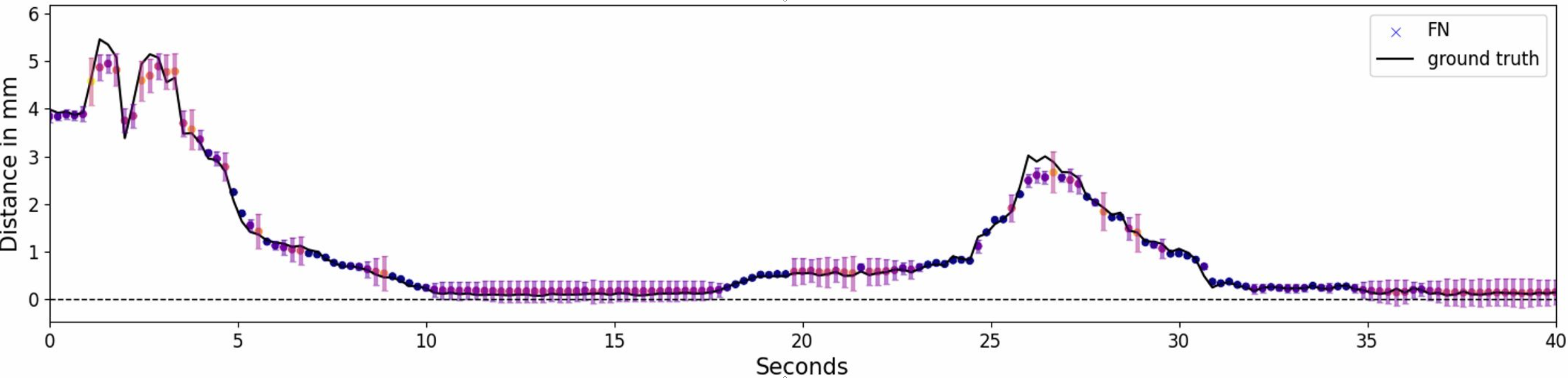}%
        \caption{Multimodality}
        \label{fig:mm_signal}
    \end{subfigure}
    \caption{Qualitative results for distance estimation on a peeling sequence using SM (a) and MM (b) models. The prediction's (points) certainty is encoded by color and error bar, where smaller and darker indicate higher certainty. A dark dot represents certainty \textgreater90\%.}
    %The ground truth (black) and predictions (color points) with error bars encoding certainty with a dot for certain predictions (\textgreater90\%).}
  \label{fig:qualitative}
\end{figure}

\bmhead{Static Multi Modality: single-modality (SM) and multimodality (MM)}
Table~\ref{tab:sm_mm_rmm} presents the comparison between the SM baseline using only OPMI images and a MM model incorporating both OPMI and iOCT data via the proposed attention-based fusion module. 
While keypoint localization accuracy showed no statistically significant difference between the SM and MM models (Welch's $t$-test, $p = 0.525$), the MM model demonstrated slightly improved instrument detection, achieving a statistically significant improvement in mAP50 over the baseline ($95.79\%$ vs $94.27\%$, $p < 0.001$, Cohen's $d = 2.99$).
    % Both models perform comparably in detection (mAP50) and keypoint localization (kp dist), indicating that OPMI imagery alone provides rich spatial cues for accurate tool tracking.
In terms of depth perception, the addition of iOCT data yields a substantial gain in accuracy (Welch's $t$-test, $p < 0.001$, Cohen's $d = 8.92$), reducing the overall mean absolute error from 480.93 µm to 128.32 µm (a 73~\% reduction). This improvement is most pronounced for close tool-tissue distances, where the error significantly decreases ($p = 0.002$) from 284.01 µm to 33.05 µm, representing an 88\added[comment={R?}]{\% reduction.}
    % In contrast, the addition of iOCT data yields a substantial gain in depth estimation accuracy, reducing the mean absolute error from 480.93 µm to 128.32 µm overall (a 73~\% reduction in error) and from 284.01~µm to 33.05~µm for close tool–tissue distances (an 88~\% reduction).

This improvement underscores the complementary nature of the modalities: OPMI captures the global surgical context while iOCT contributes localized geometric detail essential for precise depth reasoning.
Importantly, this gain comes with only minor computational overhead (22.5 ms per frame versus 18.0 ms for SM), preserving real-time performance required for intraoperative integration.

Qualitative results in Figure~\ref{fig:qualitative} illustrate the improved spatial consistency achieved through multimodal fusion.
While the SM model can infer approximate depth from visual cues such as instrument shadows, consistent with previous observations~\cite{yang2024shadowpose}, its accuracy declines when these cues are weak or absent.
The inclusion of iOCT compensates for such limitations, maintaining reliable distance estimates even when visual depth cues are ambiguous, which is particularly important for safety-critical tool–tissue interactions.

As shown in Figure~\ref{fig:certainty_vs_MAE}, the SM model also produces a calibrated certainty signal: higher predicted confidence consistently correlates with lower depth error.
For high-certainty predictions (\textgreater90 \%), the MM model significantly improves the certain dMAE from 143~µm to 54~µm compared to the SM (Welch's $t$-test, $p < 0.0001$, Cohen's $d = 4.57$), confirming that the distributional distance formulation yields a reliability measure closely aligned with prediction accuracy, which is essential for robust and trustworthy intraoperative applications.

\bmhead{Recurrent Multi Modality (RMM)}

\begin{table}[h!tb]
    \caption{Mean\,$\pm$\,std over 5 seeds for the RMM using GRU/LSTM as recurrent module on a peeling sequence with an increasing number of consecutive frames with corrupted iOCT data during inference. Masked metrics are computed only for frames with corrupted iOCT data.}
    \label{tab:rmm}
    \centering
    \begin{tabular}{crrrrr}
                \toprule
                \#corrupted iOCT & dMAE & masked dMAE & dMAE0:1 & masked dMAE0:1 \\
                frames & [$\mu m$] & [$\mu m$] & [$\mu m$] & [$\mu m$] \\
                \midrule
                \multicolumn{5}{l}{GRU}\\
                0 & 191.80 \tablestd{11.4} & - & 62.31 \tablestd{17.1} & - \\
                4 & 232.19 \tablestd{14.4} & 458.82 \tablestd{59.3} & 89.59 \tablestd{25.0} & 262.40 \tablestd{69.3}\\
                8 & 314.01 \tablestd{13.4} & 585.18 \tablestd{30.0} & 177.71 \tablestd{23.3} & 532.43 \tablestd{32.3}\\
                \midrule
                \multicolumn{5}{l}{LSTM}\\
                0 & 222.89 \tablestd{24.3} & - & 65.50 \tablestd{9.4} & - \\
                4 & 264.00 \tablestd{24.8} & 471.79 \tablestd{59.8} & 99.04 \tablestd{15.0} & 305.29 \tablestd{71.5} \\
                8 & 348.13 \tablestd{27.7} & 604.74 \tablestd{96.4} & 193.48 \tablestd{35.5} & 574.80 \tablestd{129.6} \\
                \bottomrule
    \end{tabular}
\end{table}

To assess robustness under unreliable conditions, we extend the multimodal model with the recurrent module described in Section~\ref{subsec:recurrent_module}, yielding the RMM variant.
The recurrent component integrates temporal context across 16-frame sequences and is explicitly trained to handle missing or corrupted iOCT data by randomly replacing iOCT features for selected frames during training.
As can be seen in Table~\ref{tab:sm_mm_rmm}, the recurrent models preserve strong detection (mAP50: 94.92~\tablestd{0.29}\%) and keypoint localization (kp dist: 9.65~\tablestd{0.42} px) performance, comparable to the non-recurrent models.

For evaluation, we simulate unreliable conditions by modifying iOCT embeddings for consecutive frames. The results are summarized in Table~\ref{tab:rmm} for 4-layer GRU and LSTM as recurrent model.
Because the GRU-based variant consistently outperforms the LSTM-based one, we restrict our discussion to the GRU-based model and refer to it as RMM.

For distance estimation on a single sequence, the RMM achieves a dMAE of 191.80~µm with intact iOCT inputs, but this error increases to 314.01~µm when corrupted features are introduced (584.18~µm for only corrupted frames).
While temporal fusion provides moderate resilience to short-term noise, the model fails to recover when the iOCT stream is persistently unreliable and does not reach the performance of the SM baseline in such cases.
This suggests that the network becomes over-reliant on the iOCT stream for depth perception, reflecting a form of modality collapse~\cite{chaudhuri2025modalitycollapse} commonly observed in multimodal systems.

These observations underline a broader challenge in multimodal temporal fusion: while it can enhance consistency and robustness under moderate noise, sustained reliability under varying data quality remains difficult to achieve.
Future research could therefore focus on adaptive fusion mechanisms that explicitly evaluate and balance modality contributions, fostering more robust and trustworthy multimodal behavior in intraoperative AI systems.

In our experiments, we use a synthetic dataset, as the acquisition of clinical datasets is not feasible until commercial systems become available with multimodal registration and instrument tracking for automatic OCT alignment. 
Rather than solving the Sim2Real gap, this work focuses on exploring and evaluating methodologies for multimodal fusion to demonstrate its potential and limitations.
As the current dataset does not integrate all surgical cases with complex pathologies and as real-world data can be more challenging due to varying levels of noise and more severe artifacts, the aim of this work is not to claim robustness under all intraoperative conditions, but to provide an initial, controlled analysis of the benefits and limitations of multimodal fusion, without confounding factors arising from incomplete or inconsistent annotations. 
The results of this initial study provide the community with insights into the benefits and overall challenges of multimodal OPMI-iOCT fusion, particularly in depth estimation and safety-critical tool–tissue interactions.
Fusion methodologies, combined with advancements in imaging systems, could eventually enable more comprehensive scene understanding in image-guided surgery.

\section{Conclusion}

We presented a multimodal OPMI–iOCT framework that integrates spatial, temporal, and depth information through cross-attention fusion and recurrent refinement to jointly detect, localize, and estimate the distance of surgical instruments.
Evaluated on fully annotated synthetic multimodal sequences, it establishes the first proof of principle for depth-aware, holistic scene understanding in ophthalmic surgery, demonstrating how multimodal fusion strengthens perception in safety-critical tool–tissue proximity.
By introducing a distributional distance head that provides calibrated certainty, the framework lays the groundwork for reliability-aware surgical guidance and opens pathways toward adaptive, temporally aligned multimodal fusion capable of maintaining robustness under real intraoperative conditions.

\backmatter

% \bmhead{Supplementary information}
% \textcolor{red}{Do we need to name everything included in the supplementary material here?}
% If your article has accompanying supplementary file/s please state so here. 

% Please refer to Journal-level guidance for any specific requirements.
% Supplementary materials (if any) may be submitted in addition to the manuscript file. We encourage authors to submit supplementary material along with your submission to clarify your contribution with the reviewers, which may include video files. Note that the manuscript file should stand alone, and be fully understood without the need to refer to supplementary material. Supplementary material for accepted papers will only be available online. The supplementary material should conform to the IJCARS guidelines. 

% \bmhead{Acknowledgements}
% Acknowledgements are not compulsory. Where included they should be brief. Grant or contribution numbers may be acknowledged.

% Please refer to Journal-level guidance for any specific requirements.

\section*{Declarations}
The authors declare the following competing interests: M.S. and N.N. are shareholders of SynthesEyes GmbH, which provided the synthetic datasets.

\bibliography{sn-bibliography}% common bib file
%% if required, the content of .bbl file can be included here once bbl is generated
%%\input sn-article.bbl

\end{document}